\documentclass[runningheads]{llncs}
\usepackage[T1]{fontenc}
\usepackage{algpseudocode}
\usepackage{amssymb}
\usepackage{amsmath}
\usepackage{booktabs}
\usepackage{multirow} 
\usepackage{subcaption}

\usepackage{color, colortbl}
\definecolor{LightCyan}{rgb}{0.88,1,1}

\usepackage{graphicx}
\begin{document}
\let\WriteBookmarks\relax
\def\floatpagepagefraction{1}
\def\textpagefraction{.001}

\title{D$^2$R: dual regularization loss with collaborative adversarial generation for model robustness}

\author{
Zhenyu Liu\inst{1} \and
Huizhi Liang\inst{1} \and
Rajiv Ranjan \inst{1}\and
Zhanxing Zhu\inst{2}\and
Vaclav Snasel\inst{3}\and
Varun Ojha\inst{1}
}
\authorrunning{Z. Liu et al.}

\institute{Newcastle University, Newcastle, UK \and
University of Southampton, Southampton, UK \and
Technical University of Ostrava, Ostrava, Czech Republic
}

\maketitle        
\begin{abstract}
The robustness of Deep Neural Network models is crucial for defending models against adversarial attacks. Recent defense methods have employed collaborative learning frameworks to enhance model robustness. Two key limitations of existing methods are (i) insufficient guidance of the target model via loss functions and (ii) non-collaborative adversarial generation. We, therefore, propose a dual regularization loss (D$^2$R Loss) method and a collaborative adversarial generation (CAG) strategy for adversarial training. D$^2$R loss includes two optimization steps. The adversarial distribution and clean distribution optimizations enhance the target model’s robustness by leveraging the strengths of different loss functions obtained via a suitable function space exploration to focus more precisely on the target model’s distribution. CAG generates adversarial samples using a gradient-based collaboration between guidance and target models. We conducted extensive experiments on three benchmark databases, including CIFAR-10, CIFAR-100, Tiny ImageNet, and two popular target models, WideResNet34-10 and PreActResNet18. Our results show that D$^2$R loss with CAG produces highly robust models. Our code is available at https://github.com/lusti-Yu/D2R.git.


\end{abstract}





\section{Introduction}
Deep Neural Networks (DNNs) are crucial in solving real-world problems. However, DNNs are vulnerable to adversarial attacks~\cite{szegedy2013intriguing,zhang2024improving}. Researchers have been actively investigating defense strategies to improve the robustness of DNN models against adversarial attacks. One of the most effective and well-known defense methods is adversarial training (AT)~\cite{rice2020overfitting}.  Several studies have shown that adversarial training on large models offers greater robustness (due to large learnable parameters) than the small models~\cite{schmidt2018adversarially}. Similarly, adversarial distillation (AD) methods use a pre-trained teacher model to enhance the robustness of a student model (a target model)~\cite{papernot2016distillation,goldblum2020adversarially}. Alternative to AD methods is a set of adversarial defense strategies that involves using a small model as a guide to support a target model. These small guide and target models are simultaneously trained to enhance the robustness of the target model. LBGAT is an example of a guide and target framework~\cite{cui2021learnable}. 
The advantage of using this small learnable guide model to support a target model is (a) its capacity to adaptively engage in the adversarial training process, (b) the non-requirement of a large pre-trained model as a guide, and (c) a learnable guide model can support adversarial training of any size of the target model. 

However, there are two main limitations in previous works. (i) Constraints based on a wider loss function space exploration, i.e., most works use squared error loss to implement adversarial training by leveraging the properties of a learnable guide model (e.g.,~\cite{cui2021learnable}~\cite{liu2024dynamic}). This makes AT dependent on the selected loss function and hinders it from achieving optimal performance. That is, a non-optimal choice of loss function makes insufficient guidance of the target model. (ii) Constraints based on adversarial example generation, i.e., most defense methods (such as TRADES~\cite{zhang2019theoretically} and PGD-AT~\cite{rice2020overfitting}) use a single-model and independent adversarial example generation techniques. This non-collaborative adversarial generation hinders AT, based on learnable guide-target architecture, from achieving optimal performance since it ignores collaboration between the guide and the target models.
 

\begin{figure*}[t]
\centering
\includegraphics[width=1.00\linewidth]{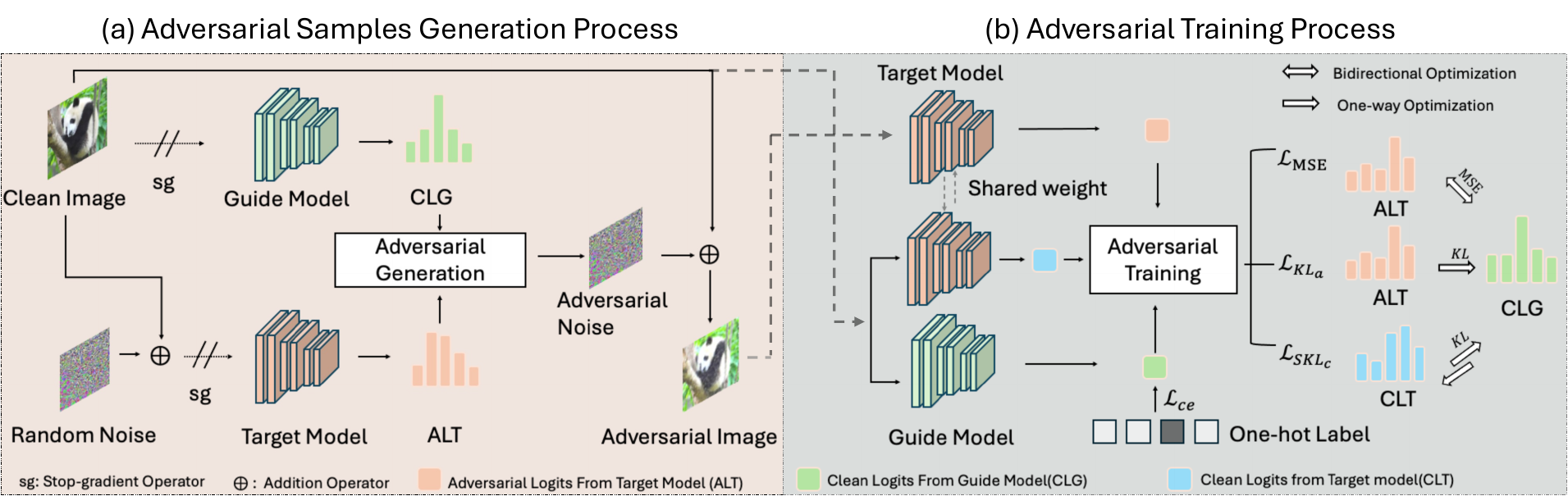}
\caption{Overview of our D$^2$R-CAG framework. (a) Proposed collaborative adversarial generation process, where the guide and target models create perturbations. These perturbations are applied to images to generate adversarial image examples. (b) D$^2$R Loss. $\mathcal{L}{ce}$ represents clean training loss on the guide model, $\mathcal{L}_{MSE}$ denotes the mean squared error between the target and guide models, $\mathcal{L}_{KL_a}$ is the optimization of adversarial distributions via KL divergence, and $\mathcal{L}_{SKL_c}$ is the optimization of clean distributions using the symmetric KL divergence function. The guide and target models are trained with synchronized inputs throughout the adversarial training process.}
\label{fig:MGLCM}
\end{figure*}

We propose a \textit{Dual Regularization Loss} (D$^2$R Loss) methodology to alleviate the limitations of loss function selection in the guide-target framework (see Fig.~\ref{fig:MGLCM}(b) that shows adversarial training process). D$^2$R loss (Fig.~\ref{fig:MGLCM}(b)) leverages the strengths of different loss functions to focus precisely on the classification distributions of guide and target models to enhance the target model's robustness. We decompose the D$^2$R loss process into two steps: (i) adversarial distribution optimization and (ii) clean distribution optimization. Specifically, a) an adversarial KL divergence ($\mathcal{L}_{KL_a}$) is employed for a precise alignment between the clean classification output distributions (logits) of the guide model and adversarial classification output distributions (logits)  of the target model; and 
b) a clean KL divergence ($\mathcal{L}_{SKL_a}$) helps with precise alignment between the clean classification output distributions  (logits) of the guide model and clean classification output distributions  (logits)  of the target model. We exploit the asymmetry of KL divergence to slightly and dynamically estimate the clean distribution gap between the guide and target models, which enhances model generalization and thereby improves robustness.


We propose a learnable \textit{Collaborative Adversarial Generation} (CAG) strategy to incorporate the adversarial example generation process into the overall adversarial training framework (see Fig.~\ref{fig:MGLCM}(a) that shows adversarial sample generation process) instead of letting the adversarial example generation happen in silos, as done in previous works. CAG strategy is based on global performance involving both guide and target model distributions. Under such adversarial example generation, we additionally leverage the learnable guide model's high confidence in its correct predictions on clean samples. As the training progresses, both the guide and target models collaboratively generate adversarial examples. This collaborative method aims to produce adaptive adversarial examples, thereby enhancing the robustness of the models. In summary, the main contribution of this work is as follows:
\begin{itemize}
\item We propose a Dual Regularization Loss (D$^2$R Loss) designed to progressively enhance the robustness of the target model. 
\item We propose a collaborative adversarial generation (CAG) method that dynamically creates adversarial examples for subsequent adversarial training. 
%
\item  We show that our method (D$^2$R Loss with CAG) outperforms other AT methods on an extensive set of experiments. 
\end{itemize}


\section{Related work}
\label{Background and Related work}

Adversarial training is the most widely used defense method. For example, Wu et al.~\cite{wu2020adversarial} proposed an effective adversarial training method called adversarial weight perturbation (AWP) to improve the DNNs' robustness. Similarly, the AT method RAT~\cite{jin2023randomized} demonstrates model robustness by adding random noise to DNN weights and a misclassification-aware adversarial training (MART)~\cite{wang2019improving} analyzes the misclassification sample to improve the robustness of DNN models. LAS-AT~\cite{jia2022adversarial} uses a single strategy model to inject an attack into a target model rather than manually crafting the attack, as is commonly done in conventional AT. TRADES~\cite{zhang2019theoretically} optimizes the trade-off between accuracy and robustness, generating adversarial examples for training through the KL-divergence loss. 

Relying on a single model for adversarial generation plays an important role in enhancing the robustness of target DNNs. However, adversarial generation using a single model suffers in achieving optimal performance in various scenarios. Therefore, our work focuses on using a learnable guide model to adaptively collaborate with the target model for generating adversarial examples. 

Our defense approach significantly differs from existing knowledge distillation methods. Traditional knowledge distillation typically employs a pre-trained guide model as a static true probability distribution and optimizes only the target model’s parameters using KL divergence. Additionally, the works that focus on using KL divergence, such as MTARD~\cite{zhao2023mitigating}, tend to only examine the knowledge scale gap between the clean teacher model and the robust teacher model. However, they do not focus on minimizing these distribution differences, and their teacher models are not learnable during training. In contrast, our research focuses uniquely on optimizing the clean probability distribution between the learnable guide model and the target model, offering a novel perspective to guide-target adversarial distillation/adversarial training frameworks. Moreover, we use small, un-trained models as guide models and train both guide and target models simultaneously for knowledge distillation. While doing so, we strengthen the loss function using our Dual Regularization Loss as opposed to other methods, such as LBGAT~\cite{cui2021learnable}, that only optimize the difference between guide and target errors using MSE.

\section{Proposed method}

\label{Proposed Method}

\subsection{Design principle}

We consider a clean dataset as $D = \{(x_{i}, y_{i})\}$ for $i = 1,2,\ldots,n$, where $x_i \in \mathbb{R}^d$ is a $d$-dimensional input and $y \in [k]$ is a class label. We have a classifier $f_\theta(x)$ that maps the inputs to the label space as $f_\theta: x \rightarrow y $ on some learnable parameters $\theta \in \mathbb{R}^p$, also called model parameters. A model parameterized by $\theta$ is attacked if for a clean input $x$ there exists an adversarial input $x^{'} = x + \delta$ such that model performance decreases, where $\delta \in \mathbb{R}^d$ is adversarial perturbation and following PGD-AT~\cite{rice2020overfitting}, $\delta$ is bounded by $l_{\infty}$. Adversarial distillation(AD)~\cite{papernot2016distillation} has demonstrated its effectiveness as a defense method to make models robust against such attacks. Typically, AD involves a pre-trained guide model. Although adversarial distillation has shown impressive performance in various studies~(e.g., \cite{hinton2015distilling}), a learnable framework based on smaller models can also achieve remarkable results (e.g.,~\cite{cui2021learnable,jia2022adversarial}). Thus, we propose a new adversarial defense method based on a learnable architecture. Specifically, traditional adversarial distillation approaches use a static, pre-trained model to transfer knowledge to the target model. However, the pre-trained model has limited adaptability. Inspired by the learnable guide model~\cite{cui2021learnable} and the pre-training teacher model~\cite{papernot2016distillation}, our goal is to further investigate the robustness advantages of the learnable guide model by exploring the gap between pre-trained and learnable methods. 

\begin{figure}[t]
    \centering
    \subfloat[\label{fig:a}]{\includegraphics[width=0.3\linewidth]{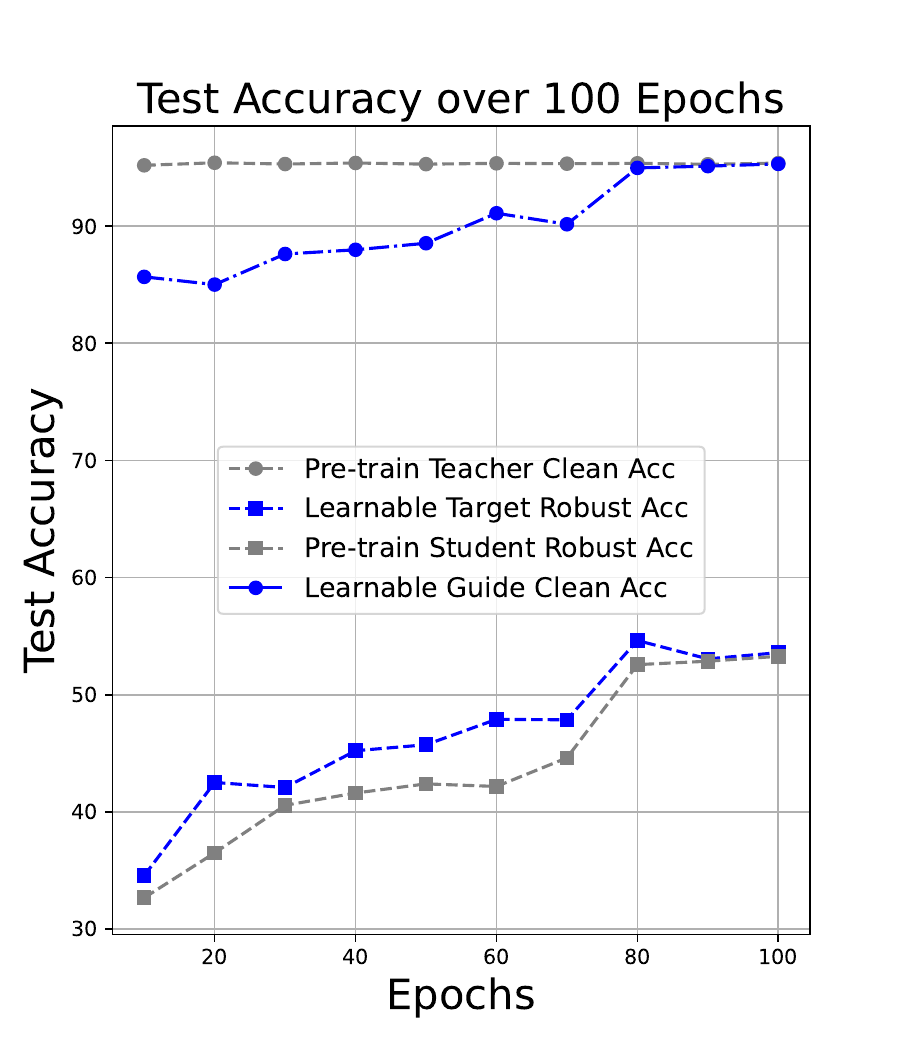}} \hfill
    \subfloat[\label{fig:b}]{\includegraphics[width=0.3\linewidth]{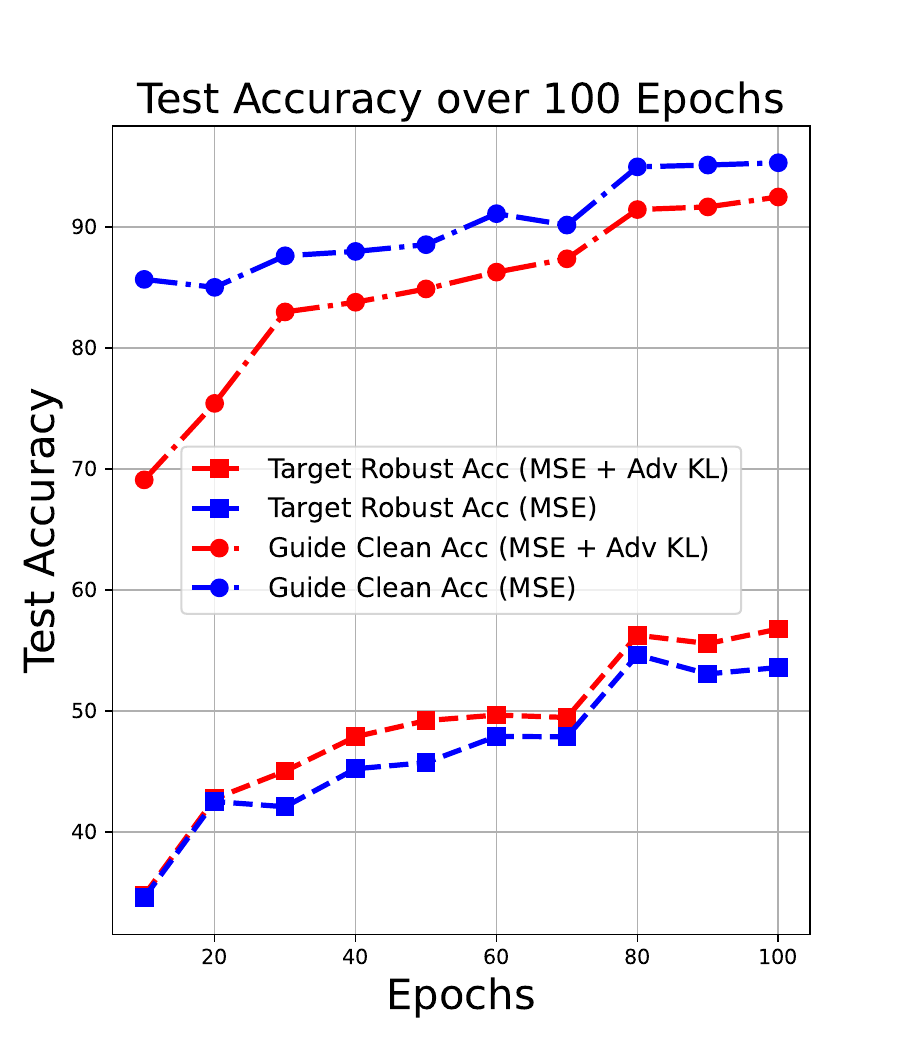}} \hfill
    \subfloat[\label{fig:c}]{\includegraphics[width=0.3\linewidth]{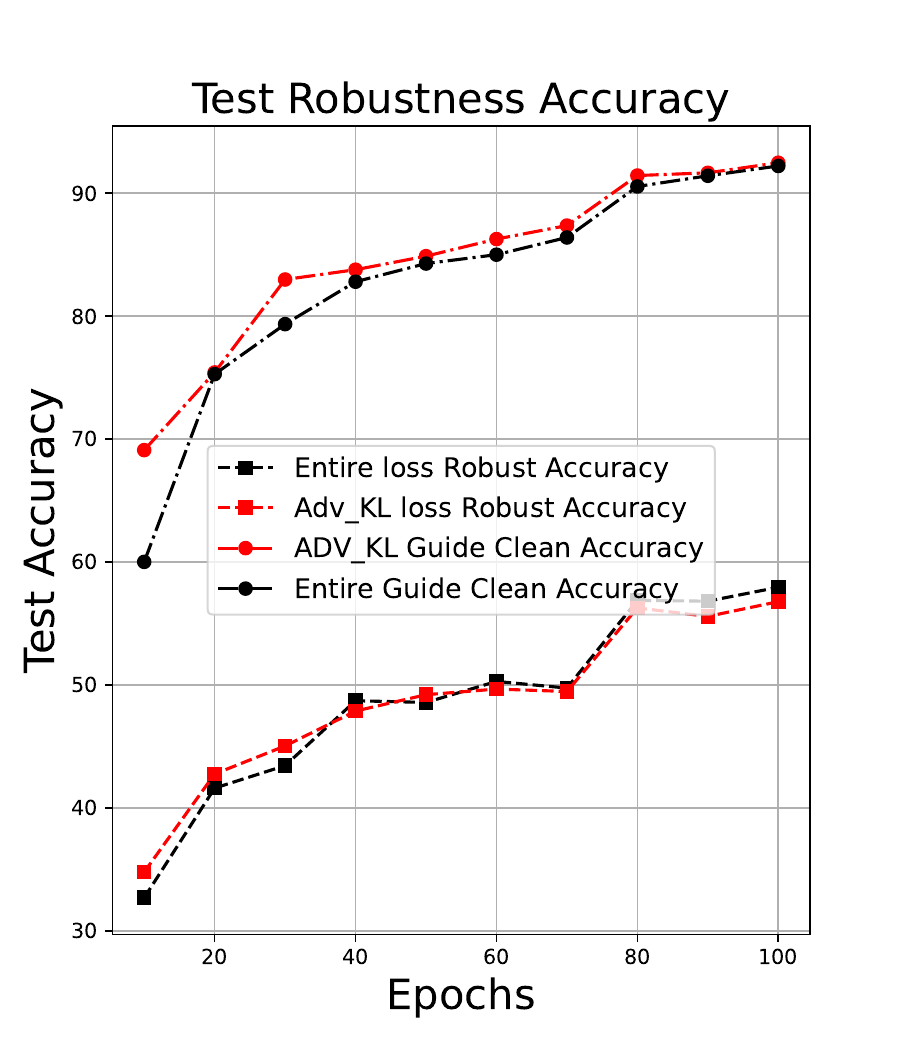}}
    \caption{Comparison of model robustness under PGD-20 attacks on CIFAR-10. 
    (a) Pre-trained ResNet-18 \textbf{(grey)} vs. learnable ResNet-18 \textbf{(blue)}. 
    (b) Guide model clean accuracy and target model robustness using our method \textbf{(red)} vs. LBGAT \textbf{(blue)}. 
    (c) Effect of Clean Symmetric Loss: without \textbf{(red)} vs. with \textbf{(black)}.}
    \label{fig:comparison}
\end{figure}

We conducted a fair comparison between the learnable guide model (ResNet-18) and the pre-trained (ResNet-18) guide model to ascertain our hypothesis. The experiment demonstrates that the learnable guide model enhances robustness more effectively (see Fig.~\ref{fig:comparison}(a)). This indicates that the learnable guide model as a teacher tends to adapt to AT better than a fixed pre-trained model as a teacher. Therefore, We conducted an experiment on the learnable guide model while using the baseline method~\cite{cui2021learnable} that uses mean square error (MSE) as the adversarial training loss. In contrast, another guide model in Fig.~\ref{fig:comparison} uses a combined loss of MSE and KL-divergence as the adversarial training loss. The experiment showed that the combined loss offered superior robustness compared to MSE alone (see Fig.~\ref{fig:comparison}(b)). Similarly, Fig.~\ref{fig:comparison}(c) shows that guide and target model output distributions on clean inputs help achieve improved performance.  
Despite the robustness improvements offered by learnable architectures, designing an effective guidance strategy is challenging due to the complex interaction between the guide and target models, which requires the target model to be much more comprehensively involved. This observation motivates us to propose a more refined and comprehensive process.

Our work performs a functional space exploration~\cite{lebedev1996introduction} aimed at enhancing robust accuracy. The loss landscape of interaction between loss functions is less understood when more than one DNNs are involved. However, 
the properties of the loss function are known: the CE measures the similarity of prediction scores (globality), MSE measures the differences between two sets of logits (locality) and KL-divergence measures the differences between two probability distributions. That is, CE showcases effectiveness in clean classification tasks, MSE loss is often used for learnable guide-target architecture, and KL is a popular choice for adversarial distillation. At the same time, the vulnerability of CE loss is well understood~\cite{pang2019rethinking}. 
The multiple loss framework harnesses the regularizations among a set of different loss functions~\cite{xu2015multi}. Since, in our framework, two DNNs interact with each other over two different tasks, the loss functions involve task-specific components for the regularization of the overall loss.       

The entire loss of D$^2$R (Dual Regularization Loss) (Fig.~\ref{fig:MGLCM}(b)) can be divided into two parts. Firstly, adversarial loss enhances the robustness of the target model by adopting KL divergence loss. Secondly, incorporating clean distribution optimization between the guide and target model can effectively improve both clean accuracy and robustness. Besides, a collaborative adversarial generation strategy (Fig.~\ref{fig:MGLCM}(a)) improves the robustness of the target model further.

\subsection{D$^2$R loss and collaborative adversarial generation}
\subsubsection{Adversarial distribution optimization} \label{Proposed advMethod} While the fundamental concept of the learnable architecture has demonstrated effectiveness, the MSE loss function still encounters precision issues that constrain its robustness. Moreover, we hypothesize that if the guide model learns the adversarial distribution of the target model more accurately,  it will lead to more synchronized and effective guidance. 

The above observation and hypothesis motivate us to propose a new method that more precisely focuses on the distribution of the target model, extending beyond the use of the MSE function to approximate the distribution of global squared differences. We propose restricting the target model's adversarial distribution based on the guide model's clean distribution. Our design incorporates operations on both MSE and KL divergence functions rather than solely using MSE because KL divergence is highly sensitive to subtle probability changes. From Fig.~\ref{fig:comparison}(b), we observe that our method is more effective at enhancing the robustness of the target model compared to relying solely on MSE. Moreover, our proposed adversarial loss enables the target model to adapt the clean distribution of the guide model. This approach enables the customization of the target model's boundaries to more precisely align with the guide model's clean status. Following the integration of adversarial optimization, both the guide model’s clean accuracy and the target model’s robustness are enhanced, as shown in Fig.~\ref{fig:comparison} (c). Consequently, the goal of enhancing robustness accuracy can be better realized by combining the KL divergence and MSE loss.

Following the analysis presented and the perspective of function properties, the KL divergence captures the subtle variances and structural characteristics of probability distributions, while MSE emphasizes the overall average error. This dual incorporation leverages KL divergence to address distributional discrepancies and MSE to quantify the overall deviation for improving robustness. Therefore, we have an effective adversarial distribution guidance (ADG) loss as:
\begin{equation}
\begin{split}
    \mathcal{L}_\text{ADG}(x,y) & = \mathop{\min}_{\theta_g, \theta_t} \mathbb{E}_{(x, y) \in D} \big\{ \mathcal{L}_{\text{CE}}(\theta_{g}, x, y) \\
    &\quad + \mathcal{L}_{\text{MSE}}(f_{g}(x), f_{t}(x^{'}))
     + \alpha \mathcal{L}_{\text{KL}}(f_{g}(x) \parallel  f_{t}(x^{'})) \big\},
\end{split}
\label{eq:advloss_function}
\end{equation}
where $\theta_g$ and $\theta_t$ are the parameters of the guide and target models; $\mathbb{E}_{(x, y) \in D}$ is the expectation of the empirical risk over the all input-output pair $(x, y) \in D$; and $\mathcal{L}_\text{CE}$, $\mathcal{L}_\text{MSE}$, and $\mathcal{L}_\text{KL}$ are the loss functions cross-entropy, mean squared error, and KL-divergence respectively. The term $f_{g}(x)$ represents the clean distribution of the guide model, while $f_{t}(x^{'})$ denotes the adversarial distribution of the target model. $ \alpha \ge 0 $ is a user-defined hyperparameter to control the strength of KL-divergence in Eq.~\eqref{eq:advloss_function}.

\subsubsection{Symmetric optimization for clean distribution}
\begin{figure}
	\begin{minipage}[t]{0.325\linewidth}
		\centering
		\includegraphics[width=1.05in]{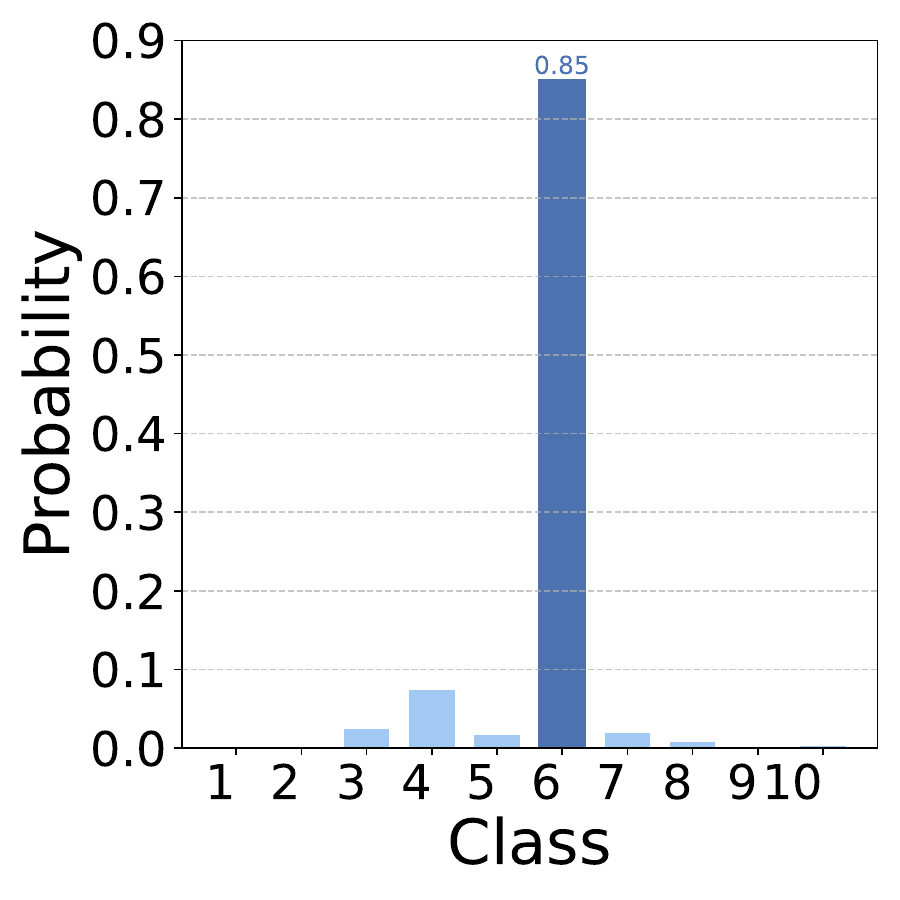}
	\end{minipage}
	\begin{minipage}[t]{0.325\linewidth}
		\centering
		\includegraphics[width=1.05in]{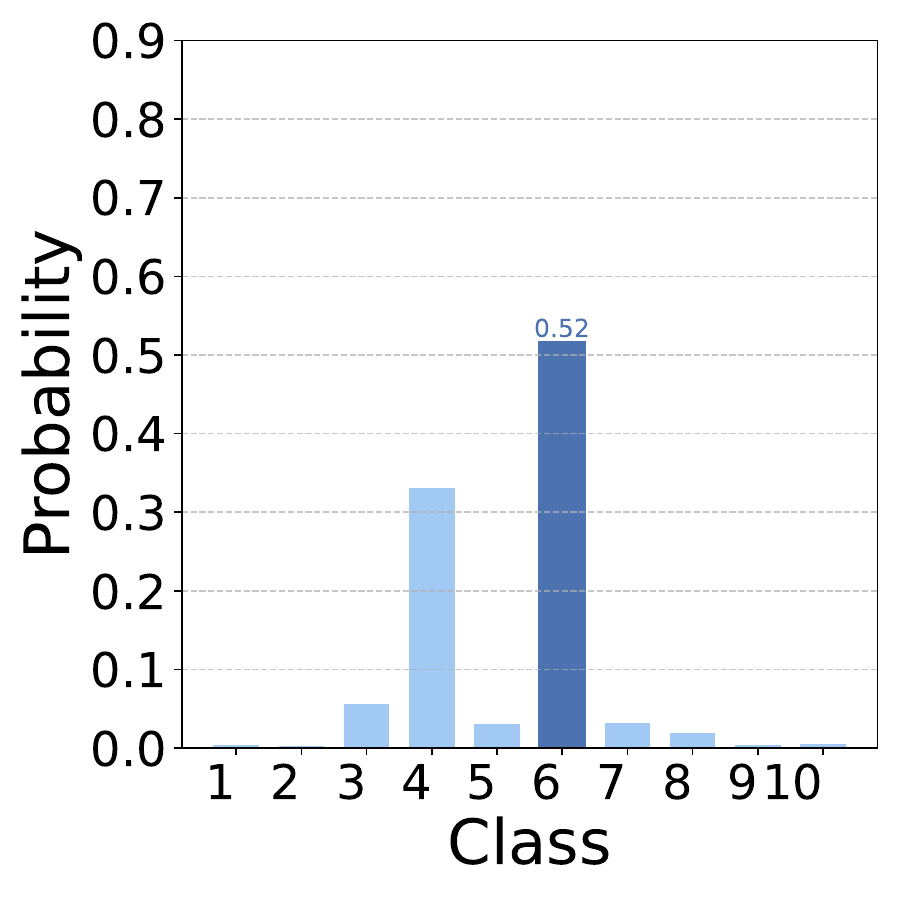}
	\end{minipage}
 	\begin{minipage}[t]{0.325\linewidth}
		\centering
		\includegraphics[width=1.05in]{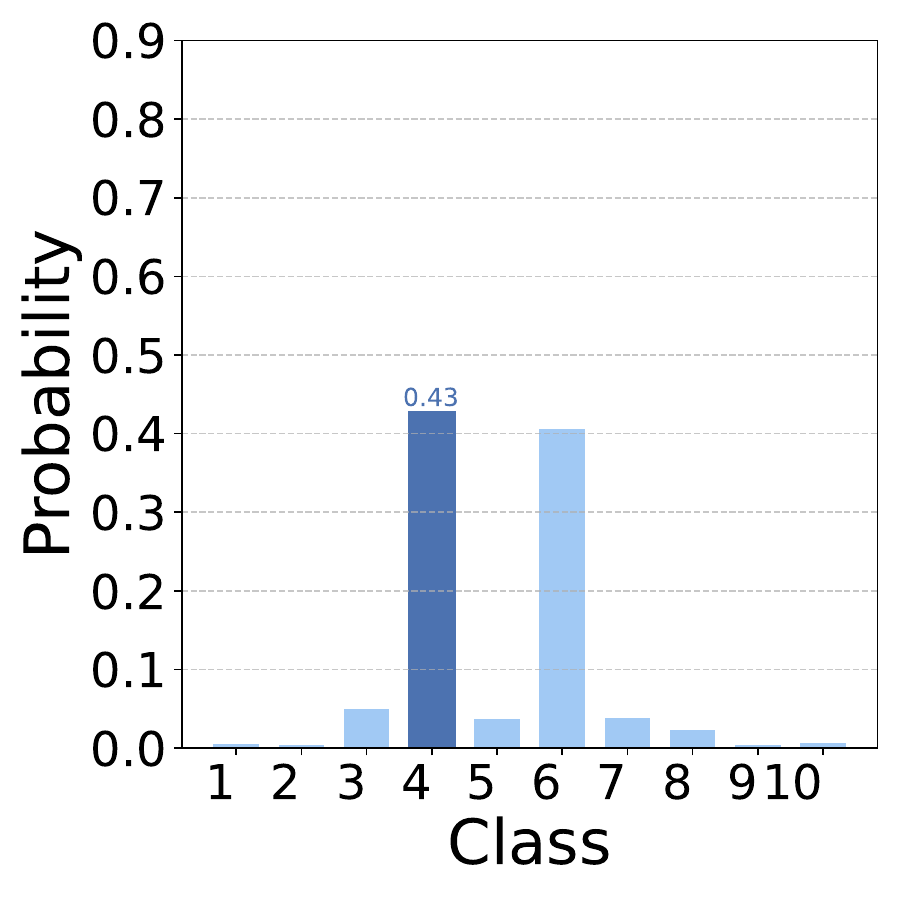}
	\end{minipage}
	\caption{An example demonstrating the effectiveness of symmetric KL. (\textbf{Left}) The probability distribution output by the guiding model is correctly classified as the 6th class (same as the ground truth). (\textbf{Center}) The probability distribution of the target model after applying the symmetric clean distribution is classified as the 6th class. (\textbf{Right}) The probability distribution of the target model without applying the symmetric clean distribution is misclassified as the 4th class.}
	\label{fig:bar_clean}
\end{figure}

To further enhance the model's robustness, we highlight the approximation of the clean outputs. During the training process of two \textit{co-trainable} models, we aim to propose an optimal approximation strategy for the clean output, ensuring robustness improvement for the target model. Although previous studies have used KL divergence for mutual learning between models, demonstrating convergence to a more robust minima with improved generalization in distillation tasks~\cite{zhang2018deep}, its application in achieving adversarial training is under-explored. In contrast, our method aims to improve the target model's robustness by dynamically optimizing the guide or target model to achieve better generalization. We adopt the difference between symmetric KL terms to optimize guide model probability distributions. Specifically, our method approximates discrepancies between different KL divergences by exploiting the positional differences of the true and target probability distributions within the KL function. We incorporate a subtle difference in KL asymmetry. As per Eq.~\eqref{eq:advloss_function} in adversarial optimization, the entire reconstruction of D$^2$R Loss is derived as:
\begin{equation} 
\begin{split}
    \mathcal{L}_\text{D$^2$R}(x,y) &= \min_{\theta_g, \theta_t} \mathbb{E}_{(x, y) \in D} \big\{ \lambda \mathcal{L}_\text{CE}(\theta_g, x, y) \\
    &\quad + \mathcal{L}_\text{MSE}(f_{g}(x), f_{t}(x^{'})) 
     + \alpha  \mathcal{L}_\text{KL}(f_{g}(x) \parallel  f_{t}(x^{'})) \\
    &\quad + \beta | \mathcal{L}_\text{KL}(f_{t}(x) \parallel  f_{g}(x)) - \mathcal{L}_\text{KL}(f_{g}(x) \parallel  f_{t}(x))| \big\},
\end{split}
\label{eq:cleanloss_function}
\end{equation}
where $\lambda \in \mathbb{R} $, $ \alpha \ge 0 $, $\beta \in \mathbb{R}$ are user-defined hyperparameters to control contributions of different loss functions. The incorporation of Eq.~\eqref{eq:cleanloss_function} effectively addresses the substantial gap between the clean distributions of the guide and target models. Specifically, based on this incorporation that this approach can effectively mitigate the target model's misclassification with generalization improvement of clean samples (as shown in Fig.~\ref{fig:bar_clean}), it further enhances the target model's robustness, as evidenced by Fig.~\ref{fig:comparison}. 

\textbf{Remark.} We denote $\mathcal{L}_\text{KL}(f_{t}(x) \parallel f_{g}(x))$ as a KL divergence quantifying the extent to which the clean distribution of the guiding model approximates the clean distribution of the target model. Conversely, $\mathcal{L}_\text{KL}(f_{g}(x)  \parallel f_{t}(x))$ quantifies the extent to which the clean distribution of the target model approximates the clean distribution of the guiding model. Therefore, we have 
\begin{equation}
\begin{aligned}
\mathcal{L}_\text{KL}(f_{t}(x)\parallel  f_{g}(x)) \ne \mathcal{L}_\text{KL}(f_{g}(x)\parallel  f_{t}(x))
\end{aligned}
\label{eq:sys}
\end{equation}

In our functional space exploration, we have two distinct optimization scenarios for the guide-target model under the clean distribution. When the value is positive, we optimize only the target model's probability distribution $f_{t}(x)$. Conversely, when the value is negative, we focus on optimizing the guide model's probability distribution 
$f_{g}(x)$. This dynamic approach allows for the flexible adjustment of either $f_{g}(x)$ or $f_{t}(x)$  based on their respective states, maintaining consistency between the guide and target clean distributions. Empirically, we observed that the value tends to oscillate slightly between positive and negative, with neither side dominating. As a result, this strategy of adapting the optimization focus between the guide and target models was empirically found to be more effective than a single, fixed optimization.

Here, we exploit the asymmetry of KL divergence to minimize the leakage of knowledge between the guide and target models by narrowing the gap between their clean probability distributions, which can improve the target model's clean and robust accuracy. Thus, we observed that when the adversarial output knowledge is sufficiently aligned, the incremental generalization on clean samples further contributes to enhancing robustness.

\subsubsection{Collaborative adversarial generation}
Adversarial generation plays a decisive role in robustness~\cite{jia2022adversarial}. In the literature, methods like TRADES~\cite{zhang2019theoretically} achieve adversarial generation with the constraint that the adversarial samples are generated using KL divergence from a single model, which falls short in mitigating the challenges of adversarial robustness.

Since adversarial generation is entangled with the expectation of adversarial gradients, it is challenging to solve such optimization problems directly based on complex frameworks. Instead, we propose an alternating adversarial generation approach, with the guide model and the target model collaboratively involved in the process. For our adversarial example generation, we solve it with the given guidance model and target model by adopting the collaboration gradient-based approach to generate the adversarial perturbation. That is, the adversarial perturbation is produced iteratively by the overall framework. We employ a guide model alongside a target model to automatically generate adversarial examples $x^{''}$. The key aspects of our methodology are as follows:
\begin{equation}
\label{eq:inner_opt}
x^{''} = \Pi  \big( \eta \, \mathrm{sign}(\nabla_{x^{'}} \mathcal{L}_\text{KL}(f_t(x^{'}) \parallel f_g(x))) + x^{'} \big),
\end{equation}
where $\mathcal{L}_\text{KL}$ represents the KL-divergence loss used in our experiment, $x^{'} = x + \delta $ is randomly selected adversarial examples, $f_{t}(x^{'})$ denotes the adversarial logits from the target model when taking $x^{'}$ as its input. Our proposed generation loss function measures the discrepancy between the adversarial target logits and the clean guide logits. It is worth noting that our collaboration adversarial generation(CAG) method integrates the Dual Regularization Loss (D$^2$R Loss) method. This integration is crucial because generating adversarial samples requires that the probability distributions of both the target model and the guide model remain within a specified range.

\section{Experiments and results}
\label{Experiments and Results}
We conducted extensive experiments on CIFAR-10, CIFAR-100, and Tiny ImageNet to evaluate our method's robust and clean accuracy under various white-box attacks. 


\textbf{Evaluation of adversarial defense.}
To validate the effectiveness of our proposed method, we compared our defense methods (D$^2$R loss without CAG and D$^2$R loss with  CAG) with several defense methods of literature and their extended versions
with respect to both robustness and clean accuracy. The evaluated methods include a range of defense methods, including 1) PGD-AT, 2) TRADES, 3) SAT, 4) MART, 5) FAT, 6) GAIRAT, (7) AWP, 8) LBGAT, and 9) LAS-AT, and 10) RAT. 

\textbf{Model settings and implementation detail.}
To ensure the comparability of results and experimental efficiency. Following by~\cite{cui2021learnable}, we adopt WideResNet34-10 as the backbone of the target network and the corresponding guide model (ResNet-18 or PreActResNet18). We set the perturbation magnitude to 0.031, the perturbation step size to 0.007, the number of iterations to 10, and the learning rate to 0.1. The batch size was set to 128.


\textbf{Comparisons on CIFAR-10 and CIFAR-100.}
The results on CIFAR-10 and CIFAR-100 are shown in Table~\ref{tb:cifar10}. We noted discrepancies between the AWP methods when combined with RAT and LAS-AT, respectively. Therefore, for a genuine and fair comparison, we have refrained from incorporating additional techniques (i.e., AWP) unless explicitly required (i.e., TRADES). Table~\ref{tb:cifar10} shows the comparison results with dataset CIFAR-10. For robust accuracy, D$^2$R performs better than LBGAT methods under all attack scenarios. Moreover, our D$^2$R-CAG method improves the performance of LBGAT under PGD-50 attack and AA attack by about 2.3$\%$ and 1.81$\%$, respectively. Furthermore, the proposed D$^2$R-CAG achieves the best robustness performance under C$\&$W and AA attack scenarios. For CIFAR100, the proposed D$^2$R-CAG achieves the best robustness performance under all attack scenarios. In detail, our D$^2$R-CAG outperforms the LBGAT 1.08$\%$ and 2.47$\%$ on the clean accuracy and PGD-50 attack accuracy, respectively. It is noteworthy that for our D$^2$R-CAG, we attribute these improvements to our use of automatically generated attack strategies, which are employed instead of relying on TRADES.

\begin{table}[h!]
\centering
\setlength{\tabcolsep}{8pt}
\caption{Robustness (\%) on CIFAR-10, CIFAR-100 with WRN34-10. TinyImgNet$^{1}$ uses PreActResNet18 as in LAS-AT. TinyImgNet$^{2}$ trains full Tiny ImageNet with WRN34-10, using ResNet18 as the guide for LBGAT and our method. The best results are in bold.}
\label{tb:cifar10}
\begin{tabular}{llcccc}
\toprule
Dataset & Method           & Clean                  & PGD-50         & C\&W           & AA             \\ \midrule
\multirow{10}{*}{\centering CIFAR-10}
& PGD-AT~\cite{rice2020overfitting}         & 85.17                  & 54.88          & 53.91          & 51.69          \\ 
& TRADES~\cite{zhang2019theoretically}          & 85.72                 & 55.9           & 53.87          & 53.40          \\ 
& MART~\cite{wang2019improving}            & 84.17                    & 58.06          & 54.58          & 51.10          \\ 
& FAT~\cite{zhang2020attacks}            & 87.97            & 48.79          & 48.65          & 47.48          \\ 
& GAIRAT~\cite{zhang2020geometry}           & 86.30                   & 58.74          & 45.57          & 40.30          \\ 
& AWP~\cite{wu2020adversarial}            & 85.57                 & 57.92          & 56.03          & 53.90          \\  
& LAS-AT~\cite{jia2022adversarial}          & 86.23 & 56.12 &  55.73  & 53.58     \\  

& RAT(TRADES)~\cite{jin2023randomized}          & 85.98                &          -     &          56.13      &  54.20  \\
& LBGAT ~\cite{cui2021learnable} (baseline)          & 88.22                &           54.30     &          54.29      & 52.23    \\ 
\rowcolor{LightCyan} & \textbf{D$^2$R-CAG (ours)} & 85.68          & 56.73       &\textbf {56.66}        & \textbf {54.65}          \\

\midrule
\multirow{8}{*}{\centering CIFAR-100}
& PGD-AT~\cite{rice2020overfitting}           & 60.89                    & 31.45          & 30.1           & 27.86          \\ 
& TRADES~\cite{zhang2019theoretically}           & 58.61                 & 28.56          & 27.05          & 25.94          \\ 
& SAT~\cite{sitawarin2021sat}              & 62.82                 &  26.76              &    27.32            & 24.57          \\ 
& AWP~\cite{wu2020adversarial}              & 60.38               & 33.65          & 31.12          & 28.86          \\ 

& LAS-AT~\cite{jia2022adversarial}          & 61.80                 & 32.54         &  31.12          & 29.03 \\

& RAT(TRADES)  ~\cite{jin2023randomized}          & 63.01                &          -     &          29.44      &  28.10  \\

& LBGAT~\cite{cui2021learnable}  (baseline)          & 60.64          &       34.62       &             30.65               & 29.33          \\ 
\rowcolor{LightCyan} & \textbf{D$^2$R-CAG (ours)}   & 61.72   & \textbf {35.01}  & \textbf {31.59}   &  \textbf {29.61}\\
\midrule
\multirow{4}{*}{\centering TinyImageNet$^{1}$}
 & PGD-AT~\cite{rice2020overfitting}     & 43.98          & 19.98          &  17.6     & 13.78          \\ 
& TRADES~\cite{zhang2019theoretically}     & 39.16          & 15.74          &   12.92    & 12.32          \\ 
& LAS-AT ~\cite{jia2022adversarial}     & 44.86          & 22.16          & 18.54 & 16.74                   \\ 

\rowcolor{LightCyan} & \textbf{D$^2$R-CAG (ours)}  & \textbf{49.42} & \textbf{22.98} & \textbf{19.29} & \textbf{16.96} \\  \midrule 
\multirow{1}{*}{\centering TinyImageNet$^{2}$}
 & LBGAT~\cite{cui2021learnable} (baseline)    & 49.11          &  24.41         & 20.22     &  18.32                  \\  

\rowcolor{LightCyan} & \textbf{D$^2$R-CAG (ours)}  & \textbf{49.47} & \textbf{24.96} & \textbf{21.10} & \textbf{19.51} \\

\bottomrule
\end{tabular}
\label{table:cifar10}

\end{table}

\textbf{Comparisons on Tiny ImageNet.}
Following~\cite{jia2022adversarial} setting, we set PreActResNet18~\cite{he2016identity} as the target model for evaluation on Tiny ImageNet. The results are shown in Table \ref{table:cifar10}. Moreover, our D$^2$R-CAG outperforms LAS-AT in all attack scenarios as well as in clean accuracy. In addition, we further provide a straightforward comparison between the LBGAT methods and D$^2$R-CAG, using the same architecture, hyper-parameters, and epoch. The results demonstrate that our method significantly outperforms LBGAT, achieving superior performance. In detail, our D$^2$R-CAG outperforms the LBGAT 1.19$\%$  AA attack accuracy.

\textbf{Ablation study}
We performed ablation experiments on the following three parameters: the two components in the D$^2$R Loss, adversarial distribution optimization loss and clean distribution optimization, and the cross-entropy (CE) loss in the guide model training. To validate the effectiveness of each component within the distribution optimization, we conducted ablation experiments on CIFAR-10 and CIFAR-100. For CIFAR-10, we trained models with varying parameters for adversarial distribution and clean distribution optimization. The trained models were then subjected to a range of adversarial attack methods. The CIFAR-10 results show that each loss function achieves optimal performance in terms of robustness. For CIFAR-100, we also experimented with different parameters for the cross-entropy (CE) loss. This indicates that the three components are compatible, and their combined use collectively enhances model robustness.

\begin{table*}[h!]

\centering

\caption{Performance on ablation study. Test robustness (\%) of model WRN34-10 on CIFAR-10. The best robustness is in bold.}

\label{tb:hpyer—cifar10}


\begin{tabular}{lcccccc}
\toprule
Method           & Clean          & PGD-10         & PGD-20         & PGD-50         & C\&W           & AA             \\ \midrule 

D$^2$R ($\alpha$ = 0, $\beta$ = 1)  & 88.35          &          55.70      &        54.28        & 53.85            &          53.82      & 52.80         \\ 

D$^2$R ($\alpha$ = 1, $\beta$ = 0) & 88.14         &          55.62       &        54.17        & 53.75           & 53.57            &  51.69        \\




D$^2$R ($\alpha$ = 20, $\beta$ = 1) & 86.48        &  57.33            & 56.23           &   55.74      & 55.02       & \textbf 53.52          \\
D$^2$R ($\alpha$ = 30, $\beta$ = 1) & 85.87        &  57.85            &56.83          & 56.50       & 55.62     & 53.87          \\
D$^2$R ($\alpha$ = 30, $\beta$ = 20) & 86.00         &  58.17        & 56.88         &  56.60       &  55.69      & 54.04         \\
D$^2$R-CAG ($\alpha$ = 30, $\beta$ = 20) & 85.68         &\textbf {58.50}          &\textbf {57.22}          & \textbf {56.73 }      &\textbf {56.66}        & \textbf {54.65} \\


\bottomrule
\end{tabular}

\end{table*}

\textbf{Sensitivity of hyper-parameter}
In our proposed algorithm, the regularization hyperparameter $\lambda$, $\alpha$ and $\beta$ is crucial. We observe that as the regularization parameter $\alpha$ increases, both robust and clean accuracy increase. The accuracy is sensitive to the regularization hyper-parameter. Considering the trade-off robustness accuracy and clean accuracy, it is not difficult to find that $\alpha = 30$ and $\beta = 20$ are suitable for outer optimization. For the parameter settings in CIFAR-10, we use $\lambda = 1.0$.  $\lambda = 0.1$ are used for cross-entropy term optimization for CIFAR-100 and Tiny ImageNet.


\section{Conclusion}
\label{Conclusion}

We have introduced D$^2$R loss, which incrementally improves model robustness through three methods. Our approach can be understood from the perspective of the impact of various loss functions on the distribution. Moreover, generating adversarial samples based on global distribution can also enhance model robustness. Finally, extensive experiments conducted on CIFAR-10, CIFAR-100, and the full Tiny ImageNet datasets demonstrate the effectiveness of our methods. 

\bibliographystyle{splncs04}
\bibliography{cas-ref}

\end{document}